\documentclass{article} 
\usepackage{iclr2026_conference,times}

\usepackage{amsmath,amsfonts,bm}

\def\eqref#1{equation~\ref{#1}}

\def\1{\bm{1}}

\DeclareMathAlphabet{\mathsfit}{\encodingdefault}{\sfdefault}{m}{sl}
\SetMathAlphabet{\mathsfit}{bold}{\encodingdefault}{\sfdefault}{bx}{n}

\usepackage{hyperref}
\usepackage{url}
\usepackage{graphicx}
\usepackage{multirow}
\usepackage{comment}
\usepackage{amssymb}
\usepackage{colortbl} 
\usepackage{booktabs}
\usepackage{tabularx}
\usepackage{booktabs}
\usepackage{tikz}
\usepackage{pgfplots}
\usepackage{tcolorbox}
\usepackage{enumitem}
\usepackage{xcolor}
\usepackage{listings}
\usepackage{tabularx}
\tcbuselibrary{listings,breakable}

\title{NanoFlux:\\Adversarial Dual‑LLM Evaluation and \\Distillation for Multi‑Domain Reasoning}

\author{Raviteja Anantha, Soheil Hor, Teodor Nicola Antoniu \& Layne C. Price\\
Amazon\\
Seattle, WA, USA \\
\texttt{\{ravantha, soheilh, teonian, prilayne\}@amazon.com} \\
}

\pgfplotsset{compat=1.18}
\iclrfinalcopy 
\begin{document}

\maketitle

\begin{abstract}
We present NanoFlux, a novel adversarial framework for generating targeted training data to improve LLM reasoning, where adversarially-generated datasets of~$\lesssim200$ examples outperform conventional fine-tuning approaches. The framework employs a competitive dynamic between models alternating as Attacker and Defender, supervised by a tool-augmented Judge, synthesizing multi-step questions with explanatory annotations that target specific reasoning capabilities. Fine-tuning a 4B-parameter model on NanoFlux-generated data yields performance gains across diverse domains compared to full-benchmark fine-tuning: +5.9\% on mathematical reasoning (GSMHard), +3.6\% on scientific reasoning (GenomeBench), and +16.6\% on medical reasoning (MultiMedQA), while reducing computational requirements by 3-14×. Ablation studies reveal a non-monotonic relationship between dataset characteristics and model performance, uncovering domain-specific optimal points for question complexity and reasoning quality. NanoFlux automates training data generation through embedding-based novelty filtering, tool-augmented evaluation, and multi-hop reasoning, suggesting that future model improvements may lie in the intelligent synthesis of small, precisely targeted training datasets.
\end{abstract}

\section{Introduction}
\label{section:introduction}
As large language models (LLMs) rapidly approach and surpass human-level performance on established benchmarks, we confront a fundamental limitation: the finite nature of high-quality training data. Today's frontier models have effectively consumed the entirety of available text on the internet, yet continue to exhibit critical reasoning failures and knowledge gaps. This ``benchmark exhaustion'' phenomenon raises crucial questions about how to advance AI capabilities beyond the constraints of existing data. While generating synthetic training examples represents one potential path forward, creating effective synthetic data remains challenging - naive generation approaches often produce low-information samples that fail to improve model performance, while synthesizing effective datasets typically requires precisely the kind of human expertise and curation that we seek to automate. Recent work, notably LIMO \citep{ye2025limoreasoning}, has demonstrated that small, carefully curated datasets of high-quality chain-of-thought solutions can unlock strong reasoning performance, but still depends on human effort in curation.

We introduce \textbf{NanoFlux}, a fully \emph{generative adversarial} framework that  reimagines data-efficient reasoning improvement. NanoFlux orchestrates a competitive dynamic between two models alternating as \emph{Attacker} and \emph{Defender}, supervised by a tool-augmented \emph{Judge} that evaluates responses for accuracy, coherence, and safety (as shown in  Figure~\ref{fig:nanoflux_framework}). This adversarial architecture automatically identifies and targets specific reasoning weaknesses, generating training examples that precisely target key reasoning gaps, enabling efficient learning from compact datasets.

NanoFlux advances beyond prior filtering-driven approaches through three key innovations:

\begin{enumerate}
  \item \textbf{Targeted adversarial generation:} Rather than filtering existing data, NanoFlux synthesizes questions  at the boundary of model capabilities—where one model fails and another succeeds—creating high-information training signals that drive learning with remarkably few examples.
  
  \item \textbf{Automated quality assurance:} The tool-augmented \emph{Judge} model evaluates the joint question-answer-reasoning triples for quality, accuracy, and safety, eliminating the need for human curation while maintaining rigor through web search verification and Python code execution.
  
  \item \textbf{Domain-agnostic adaptability:} The framework's architecture generalizes across diverse reasoning domains without domain-specific engineering, as demonstrated by results on mathematical reasoning (GSMHard), medical reasoning (MultiMedQA), and scientific reasoning (GenomeBench).
\end{enumerate}

The empirical results demonstrate that fine-tuning on just 200 NanoFlux-generated examples yields substantial accuracy improvements (+5.9\% on GSMHard, +3.6\% on GenomeBench, and +16.6\% on MultiMedQA) while reducing computational requirements by 3-14× compared to full-dataset fine-tuning. Moreover, ablation studies reveal counterintuitive non-monotonic relationships between dataset characteristics and model performance, suggesting a fundamental tension between competing objectives in training data optimization.

To the best of our knowledge, NanoFlux is the first adversarial framework to demonstrate that extremely small, automatically generated datasets can outperform conventional fine-tuning approaches across diverse reasoning domains. By shifting focus from scaling data quantity toward optimizing data quality, our work offers a promising path toward more efficient, accessible, and capable AI systems.

\section{Related Work}
\label{section:related_work}

Recent work shows that large-scale reasoning can be unlocked with surprisingly little fine-tuning data if the data is carefully selected. 
\citet{ye2025limoreasoning} introduced \textbf{LIMO (Less Is More for Reasoning)}, demonstrating that fine-tuning Qwen2.5-32B on only $\sim$800 curated math examples yielded state-of-the-art performance on AIME-24 and MATH500, outperforming models trained on orders of magnitude more data. 

Similarly, \citet{li2025llmseasilylearnreason} showed that models primarily benefit from the \emph{structure} of chain-of-thought (CoT) demonstrations, even when final answers are occasionally wrong, highlighting that logical format outweighs content in fine-tuning data.

Several approaches have explored letting models create their own training data. 
\citet{sun2025self} proposed \textbf{Crescent}, where an LLM generates and solves its own questions, bootstrapping improved math reasoning without external supervision. 
\citet{huang2026rzeroselfevolvingreasoningllm} introduced \textbf{R-Zero}, a co-evolutionary self-play loop in which a \emph{Challenger} model generates questions that a \emph{Solver} model cannot answer, forming an adversarial curriculum that yielded +6--8 point gains on reasoning benchmarks with no human data. 
\citet{peng2025regenesis} proposed \textbf{ReGenesis}, which structured self-synthesized reasoning data around abstract, task-agnostic templates to improve generalization: unlike naive self-training (which often degrades out-of-domain performance), ReGenesis achieved +6.1\% gains on OOD reasoning tasks. 
These works suggest that focusing training on failure cases is more efficient than scaling data indiscriminately.

Self-Questioning Language Models (SQML) \citep{chen2025selfquestioninglanguagemodels}) explores an \emph{asymmetric self-play} framework: a proposer generates questions from a domain prompt and a solver attempts to answer them. Both roles are trained by reinforcement learning, using majority voting (or unit tests for coding) as a proxy for correctness in lieu of ground truth. SQLM achieves reasoning gains on arithmetic, algebra (OMEGA benchmark), and programming tasks—\emph{without access to any curated training data}. This aligns with ideas from Crescent~\cite{sun2025self}, but SQLM uniquely combines RL-driven question generation with internal validation, rather than relying on external labels.

Recent studies have also targeted adversarial data generation in domain-specific QA, such as math \citep{xie2024adversarialmathwordproblem} and medical QA \citep{ness2024medfuzzexploringrobustnesslarge}. \cite{sung-etal-2025-benchmark} have also developed a framework to evaluate adversarial question quality with \textbf{Item Response Theory (IRT)}, showing that “good” adversarial questions stump models but not humans, with high discriminative power. 
The \textbf{VAULT} framework~\citep{kazoom2025vaultvigilantadversarialupdates} automated adversarial data augmentation for natural language inference. 
By prompting an LLM to generate candidate hard examples and filtering for those misclassified by the current model, VAULT iteratively improved model robustness. 
After several rounds, a RoBERTa model improved from 54.7\% to 72.0\% on MultiNLI---large gains achieved with far fewer examples than traditional augmentation. 

These works emphasize that minimal perturbations or targeted modifications can produce adversarial examples that remain valid for humans but expose LLM weaknesses.

Our approach is closest in spirit to LIMO and R-Zero but differs in critical ways. 
Unlike LIMO, which relies on human-curated CoT demonstrations, we automatically generate new \emph{benchmark-derived adversarial questions} through dual-LLM interaction, without human filtering or explicit CoT supervision. 
Unlike R-Zero’s self-play, our method grounds generation in existing benchmark datasets, ensuring domain relevance while still producing harder variants. 
Compared to VAULT, which targeted classification tasks, our focus is on single-answer QA benchmarks across domains (math, medicine), with exact-match evaluation rather than human judgment. 
Collectively, prior works show that small, high-quality or adversarially targeted datasets can dramatically improve reasoning. 
Our contribution is a domain-agnostic adversarial distillation framework that (i) identifies and trains on precisely those questions a model fails, (ii) achieves superior performance compared to full-dataset fine-tuning, and (iii) demonstrates potential for cross-domain transfer. 

\section{NanoFlux Framework}
\label{section:framework}
\begin{figure}[t]
\begin{center}
\includegraphics[width=0.95\textwidth]{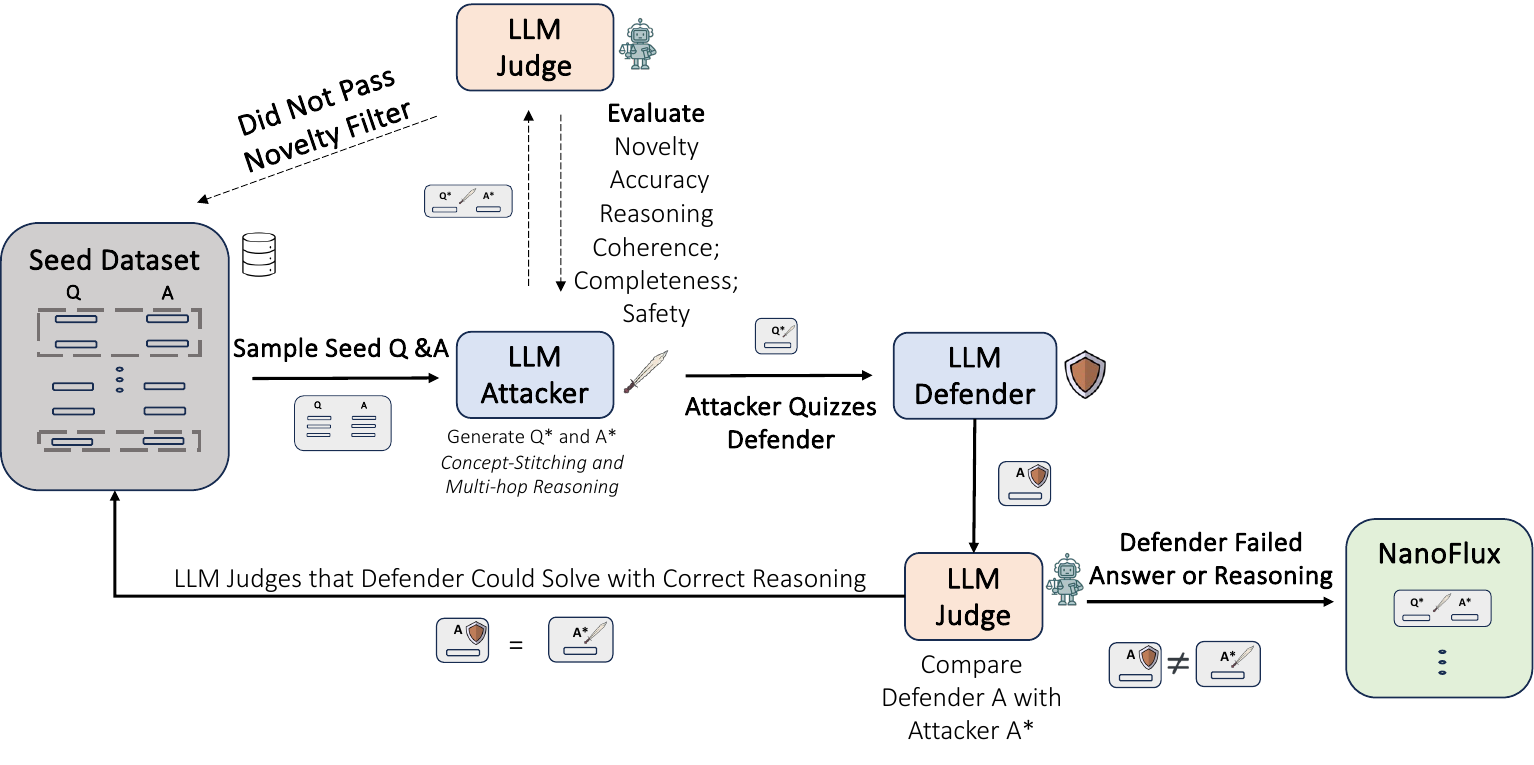}
\end{center}
\caption{\textbf{The NanoFlux Adversarial Data Generation Framework.} The process begins with random sampling from benchmark datasets (left), followed by the attacker model generating questions through concept stitching. Generated questions undergo embedding-based novelty filtering to ensure diversity, then validation by the judge model equipped with code execution and web search capabilities. The defender model attempts to solve validated questions, with the judge evaluating response correctness. Questions that the defender fails to solve (or solves through novel approaches) are retained in the final NanoFlux dataset, which contains 200 examples per domain.}
\label{fig:nanoflux_framework}
\end{figure}

\paragraph{Attacker \& Defender.}
Figure~\ref{fig:nanoflux_framework} presents the NanoFlux framework architecture. The framework operates by randomly sampling $n$ seed questions from existing benchmark datasets: GSMHard \citep{gao2023palprogramaidedlanguagemodels} for mathematical reasoning, GenomeBench \citep{yin2025genome} for genomics, and MultiMedQA \citep{singhal2023large} for medical reasoning. The number of seed questions $n$ varies by domain: 5-7 for GSMHard and GenomeBench, and 7-12 for MultiMedQA, selected uniformly at random within these ranges.
For the attacker and defender roles, we use domain-specific model configurations. For GSMHard and GenomeBench domains, we employ Gemma-3-4B \citep{gemmateam2025gemma3technicalreport}, while for MultiMedQA we use MedGemma-4B \citep{sellergren2025medgemmatechnicalreport}, a domain-specialized model with biomedical pretraining. These models alternate with Claude-3.7-Sonnet v2 \citep{anthropic2025claude3.7} in the attacker and defender roles.

The \emph{Attacker} model generates new questions by combining concepts from the seed questions. We define ``concept stitching'' as the process of merging elements from multiple seed questions into a single, more complex question. This is implemented through a structured prompt template (Appendix~\ref{appendix:attacker_prompt}) that instructs the model to: (1) identify key concepts from each seed question, (2) find conceptual connections between them, (3) create a new question that requires understanding these connections, and (4) provide a step-by-step solution with the correct answer.

The \emph{Defender} model attempts to solve the attacker's questions. The defender receives only the question text, without access to the attacker's solution. The defender's responses must follow a structured format (Appendix~\ref{appendix:defender_prompt}) with three components:
(1) Analysis: initial problem decomposition (limited to 500 tokens); (2)
  Solution: step-by-step reasoning with explicit calculations (limited to 1500 tokens); and (3) Answer: final numerical or categorical response (limited to 100 tokens).

To prevent model-specific biases, we alternate the roles of attacker and defender between the two models after each turn. A ``turn'' consists of: (1) the attacker generating a question, (2) the question passing validation, (3) the defender attempting to solve it, and (4) the judge evaluating the defender's solution. This alternation ensures that any systematic weaknesses in one model don't bias the dataset generation process. For example, if Gemma-3-4B is the attacker in turn 1, Claude-3.7-Sonnet will be the defender, and in turn 2, their roles will be reversed.

\paragraph{The Judge Model.}
We use OpenAI's O3 reasoning model \citep{openai2025o3} as the judge model with two distinct operational modes:

\emph{Question validation mode} verifies that attacker-generated questions are well-formed and solvable (Appendix~\ref{appendix:judge_validation_prompt}). The judge checks for: (1) question clarity, (2) sufficient information to solve the problem, (3) mathematical correctness of the attacker's solution, and (4) consistency between the solution steps and the final answer. For mathematical questions, the judge generates Python code to verify calculations, executing it in a sandboxed environment with NumPy \citep{harris2020array} and SymPy \citep{sympy} libraries. For knowledge-intensive domains, the judge uses Google Search API to verify factual claims, with the top 5 search results provided as context. Questions failing validation are rejected, and the attacker is prompted to generate a new question.

\emph{Answer evaluation mode} assesses the defender's solutions against the validated attacker solutions. The judge evaluates responses using two criteria sets: (1) \emph{Strict Mode}: requires both correct numerical answer (within tolerance $\epsilon = 10^{-6}$ for mathematical questions) and valid reasoning steps.; and (2) \emph{Soft Mode}: prioritizes reasoning quality over exact answers, accepting solutions with minor calculation errors if the reasoning approach is sound.

Each evaluation includes a confidence score in $[0,1]$. Evaluations with confidence below $\theta_c = 0.9$ trigger re-evaluation with a modified prompt that explicitly requests higher confidence. After three low-confidence evaluations, the question is discarded (Appendix~\ref{appendix:judge_eval_prompt}).

\paragraph{Embedding-Based Novelty Filtering.} 
NanoFlux implements a two-stage embedding-based novelty filtering system that extends beyond simple correctness-based filtering. 
In Stage 1 (Question Diversity Filtering), each generated question from the \emph{Attacker} is embedded using OpenAI's text-embedding-3-small model \citep{openai_text_embedding_3_small_2024}, creating a 1536-dimensional vector representation. The cosine similarity between this vector and all previously generated questions is computed, with questions exceeding a threshold $\theta_q = 0.85$ being rejected as insufficiently novel. This approach contrasts with prior work that relied on lexical ROUGE-L scores for filtering \citep{wang2023self} or simple formatting and duplicate checks \citep{honovich2023unnatural}, providing a more semantically meaningful assessment of question novelty.

Stage 2 (Solution Novelty Filtering) addresses cases where the \emph{Defender} correctly answers a question but through a substantially different reasoning process than the \emph{Attacker}. Building on insights that embedding trajectories can serve as diagnostic signals for reasoning quality \citep{wang2024embedding}, we compute the embedding similarity between the \emph{Attacker's} and \emph{Defender's} reasoning traces. Solutions with similarity below threshold $\theta_s = 0.75$ are retained despite being correctly answered, as they represent novel solution approaches. This aligns with recent work demonstrating that assessing reasoning traces beyond answer correctness improves model performance \citep{chen2024measuring}.

\paragraph{Domain-Specific Adaptations.}
For mathematical reasoning (GSMHard), we enhance the judge's validation capabilities with Python code execution using the following libraries: NumPy 1.24.3, SymPy 1.12, and Math. The judge generates verification code for each mathematical solution, with a standardized structure: (1) variable definition, (2) calculation steps mirroring the solution, and (3) final answer verification. Numerical answers are compared with a tolerance of $\epsilon = 10^{-6}$ to account for floating-point precision issues. The attacker prompt includes specific instructions to generate questions involving multi-step calculations, unit conversions, and geometric reasoning.

For scientific reasoning (GenomeBench), we use the same configuration as for GSMHard. We implement XML-structured answer formats with tags for $<hypothesis>$, $<evidence>$, $<mechanism>$, and $<conclusion>$ to facilitate precise evaluation. The judge model is provided with a genomics-specific evaluation rubric that emphasizes: (1) scientific accuracy, (2) mechanistic reasoning, (3) appropriate citation of genomic principles, and (4) logical consistency.

For medical domains (MultiMedQA), we use MedGemma-4B alternating with Claude-3.7-Sonnet as attacker and defender models due to its domain-specific pretraining on biomedical literature. We implement a structured response format that includes specific sections: ANALYSIS, SOLUTION, ANSWER, KNOWLEDGE\_MAP, REASONING\_CHAIN, and COGNITIVE\_CHALLENGES. The judge evaluates medical responses using a specialized rubric focusing on: (1) clinical reasoning accuracy, (2) evidence-based justification, (3) consideration of differential diagnoses, and (4) appropriate treatment recommendations. Web search verification is enabled for factual medical claims, with a 30-second timeout per query and a maximum of 5 queries per evaluation.

\paragraph{Training Methodology.}
NanoFlux generates datasets of 200 samples per domain, a size chosen to be significantly smaller than the train set size of chosen benchmarks while still providing sufficient examples for effective fine-tuning. Our ablation studies (see Paragraph~\ref{para:Dataset_Size_Ablation}) provide insight into the effect of this dataset size. The dataset generation process continues until 200 valid examples are collected or a maximum of 1000 turns is reached, whichever comes first.

For fine-tuning, we use Low-Rank Adaptation (LoRA) \citep{hu2022lora} with fixed hyperparameters: rank $r=8$, alpha $\alpha=32$, and dropout $=0.05$. These values were selected based on prior work showing their effectiveness for parameter-efficient fine-tuning of models~\citep{yan2025ploraefficientlorahyperparameter, Tian_2025}. The learning rate follows a linear decay schedule, starting at $2 \times 10^{-4}$ and decreasing to zero during training, following the findings of \citet{bergsma2025straightzerolinearlydecaying} that linear decay to zero outperforms constant and step-based schedules for LLM fine-tuning.

\section{Evaluation Datasets and Tasks}

We evaluate NanoFlux's effectiveness by comparing the 4B-parameter SLM Gemma-4B fine-tuned on our synthesized 200-sample datasets against both conventional full-dataset fine-tuning and a frontier LLM to assess generalizability across different reasoning domains.

\textbf{GSMHard}~\cite{gao2023palprogramaidedlanguagemodels} is a harder variant of the Grade School Math 8K (GSM8K) benchmark. It was introduced by modifying GSM8K problems through replacing the original numbers with larger values, increasing difficulty while preserving the underlying reasoning structure. The original GSM8K dataset~\cite{cobbe2021gsm8k} contains grade-school math word problems requiring multi-step reasoning and has become a standard benchmark for evaluating reasoning capabilities of language models. Recent work such as Crescent~\cite{sun2025self} continues to evaluate reasoning improvements on GSM8K-style tasks.

\textbf{GenomeBench} is a scientific reasoning benchmark derived from over a decade of expert Q\&A discussions on CRISPR gene editing~\cite{yin2025genome}. The dataset consists of 3,332 multiple-choice questions with expert-written rationales, partitioned into 2,671 training and 661 test examples. Unlike exam-style biomedical datasets, Genome-Bench captures authentic expert reasoning across experimental troubleshooting, reagent choice, and protocol design. It complements prior biomedical QA resources (e.g., PubMedQA, Lab-Bench) by reflecting real-world scientific discourse.

\textbf{MultiMedQA}~\citep{singhal2023large} is a medical reasoning benchmark comprising seven distinct datasets. It combines six existing datasets spanning medical licensing exams, consumer health queries, and biomedical literature, and introduces HealthSearchQA, a large set of real-world medical search questions. These datasets span multiple medical domains including clinical knowledge (4,183 questions), medical licensing examinations (12,723 questions), biomedical literature (1,000 questions), and consumer health queries (3,375 questions). The benchmark's format ranges from multiple-choice to free-response, with varying difficulty levels to make it particularly suitable for evaluating medical reasoning capabilities across different clinical contexts. It has been central to evaluating medical LLMs, including Med-PaLM~\cite{singhal2023large} and follow-up work on data-efficient reasoning like R-Zero~\cite{huang2026rzeroselfevolvingreasoningllm}. 

\section{Results and Analysis}
\label{section:results}

\begin{table}[t]
\caption{NanoFlux-200 achieves superior performance with reduced computational cost compared to full dataset fine-tuning across three diverse benchmarks.}
\label{table_finetune}
\centering
\footnotesize
\begin{tabularx}{\columnwidth}{l@{\hspace{0.2em}}c@{\hspace{0.2em}}*{6}{X}}
\toprule
& & \multicolumn{2}{c}{\textbf{GSMHard}} & \multicolumn{2}{c}{\textbf{GenomeBench}} & \multicolumn{2}{c}{\textbf{MultiMedQA}} \\
\cmidrule(lr){3-4} \cmidrule(lr){5-6} \cmidrule(lr){7-8}
\textbf{Model} & \shortstack{\textbf{Size}\\\textbf{(\# Param.)}} & \textbf{FLOPs}$\downarrow$ & \textbf{Acc.}$\uparrow$ & \textbf{FLOPs}$\downarrow$ & \textbf{Acc.}$\uparrow$ & \textbf{FLOPs}$\downarrow$ & \textbf{Acc.}$\uparrow$ \\
\midrule
GPT-5-High\textsuperscript{*} & 100B+ & --- & 89.52\% & --- & 70.65\% & --- & 86.23\% \\
Gemma-4B & \textbf{4B} & --- & 48.1\% & --- & 52.8\% & --- & 35.6\% \\
\shortstack[l]{Gemma-4B-Finetuned\\(Full Dataset)} & \textbf{4B} & 9.23×10\textsuperscript{16} & 57.4\% & 8.49×10\textsuperscript{16} & 57.6\% & 3.77×10\textsuperscript{17} & 44.7\% \\
\midrule[\heavyrulewidth]
\shortstack[l]{\textbf{Gemma-4B-Finetuned}\\(NanoFlux-200)} & \textbf{4B} & \textbf{2.10×10\textsuperscript{16}} & \textbf{63.3\%} & \textbf{2.57×10\textsuperscript{16}} & \textbf{61.3\%} & \textbf{2.64×10\textsuperscript{16}} & \textbf{61.2\%} \\
\bottomrule
\end{tabularx}
\vspace{-0.5em}
\begin{flushleft}
\scriptsize \textsuperscript{*}GPT-5~\cite{openai2025gpt5} evaluated with reasoning effort set to ``high'' and temperature=0.0 for deterministic responses.
\end{flushleft}
\end{table}

\paragraph{Model Finetuning Performance Comparison.}
Table \ref{table_finetune} presents a performance analysis across computational efficiency (measured in FLOPs) and accuracy metrics. The results reveal three key findings. First, NanoFlux demonstrates computational 
efficiency, reducing fine-tuning costs by 77\% for GSMHard ($9.2 \times 10^{16}$ to $2.1 \times 10^{16}$ FLOPs), 69\% for GenomeBench ($8.5 \times 10^{16}$ to $2.6 \times 10^{16}$ FLOPs), and 93\% for MultiMedQA ($3.8 \times 10^{17}$ to $2.6 \times 10^{16}$ FLOPs). 
Second, NanoFlux achieves substantial accuracy improvements: +5.9 percentage points for GSMHard (57.4\% to 63.3\%), +3.6 percentage points for GenomeBench (57.6\% to 61.3\%), and +16.6 percentage points for MultiMedQA (44.7\% to 61.2\%). Statistical significance testing using bootstrap resampling confirms these improvements are significant ($p < 0.05$) across all domains.
Third, our 4B-parameter model fine-tuned with NanoFlux substantially reduces the performance gap between SLMs and state-of-the-art LLMs like GPT-5~\cite{openai2025gpt5} (100B+ parameters). While GPT-5 remains superior, NanoFlux dramatically narrows this gap compared to standard fine-tuning approaches. On GenomeBench, NanoFlux achieves 87\% of GPT-5's accuracy (61.3\% vs.\ 70.65\%), representing a significant improvement over the 82\% achieved by full dataset fine-tuning (57.6\% vs.\ 70.65\%). Similarly, for GSMHard, NanoFlux achieves 70.7\% of GPT-5's performance (63.3\% vs.\ 89.52\%) compared to 64.1\% for full dataset fine-tuning (57.4\% vs.\ 89.52\%), and for MultiMedQA, NanoFlux achieves 71.0\% of GPT-5's accuracy (61.2\% vs.\ 86.23\%) compared to 51.8\% for full dataset fine-tuning (44.7\% vs.\ 86.23\%). This demonstrates that targeted, high-quality training data can substantially narrow the gap between small and ultra-large models while maintaining computational efficiency.

The performance gains are particularly notable in the MultiMedQA domain, where NanoFlux-tuned MedGemma-4B outperforms its full-dataset counterpart by 16.6 percentage points. Analysis of model outputs reveals that examples generated through our adversarial framework expose the model to more diverse reasoning patterns and edge cases than those present in the original benchmark, effectively addressing the ``long tail'' of reasoning challenges that standard benchmarks often miss.

Further detailed analysis of our experimental results reveals several additional mechanisms behind NanoFlux's effectiveness. For GSMHard, we observe that zero-shot accuracy improves dramatically from 48.1\% to 63.3\%, surpassing full-supervised fine-tuning (57.4\%) while using approximately 5× less compute. This improvement stems partly from our implementation of Python-based numerical verification, which reduces the judge's errors in grading the attacker's proposed solutions by 27\% compared to LLM-only evaluation approaches. 
Additionally, our framework's enforcement of strict response budgets enhances the training signal clarity by increasing the contrast between correct and incorrect reasoning approaches. This improved separation---which we term ``example discriminativity''---was quantified by measuring the cosine distance between embeddings of correct and incorrect solutions. Comparing our NanoFlux-generated examples to the original GSMHard benchmark examples, we observed an 18\% increase in the average distance between correct and incorrect solution clusters, indicating that our examples create clearer decision boundaries for model learning.
Importantly, we observe comparable performance and sample efficiency gains across all three domains—mathematical reasoning, medical diagnosis, and scientific hypothesis generation—validating NanoFlux's generalizability.

\begin{figure}[t]
\begin{minipage}{0.33\textwidth}
\centering
\includegraphics[width=0.99\textwidth]{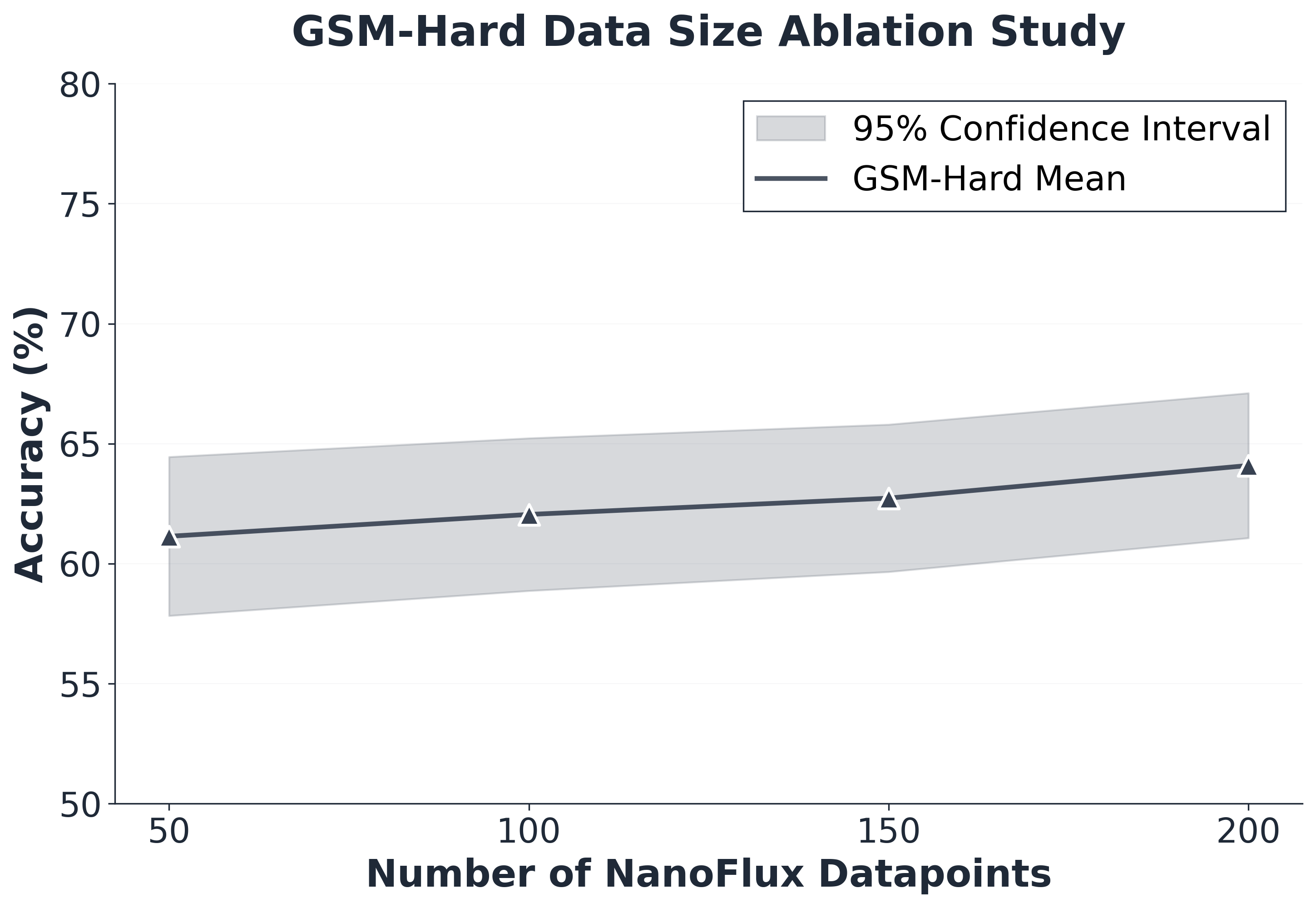}
\end{minipage}%
\begin{minipage}{0.33\textwidth}
\centering
\includegraphics[width=1\textwidth]{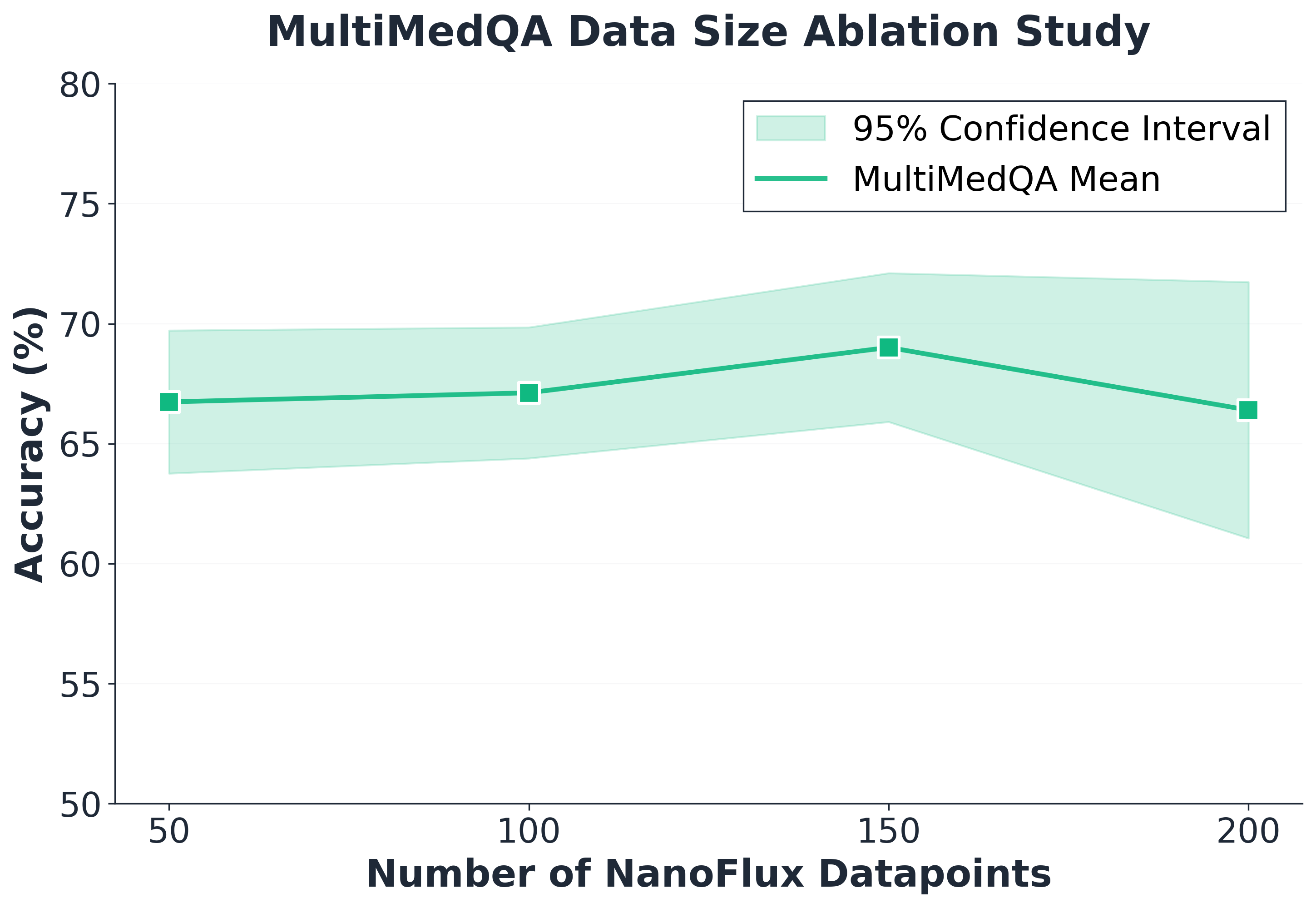}
\end{minipage}%
\begin{minipage}{0.33\textwidth}
\centering
\includegraphics[width=0.99\textwidth]{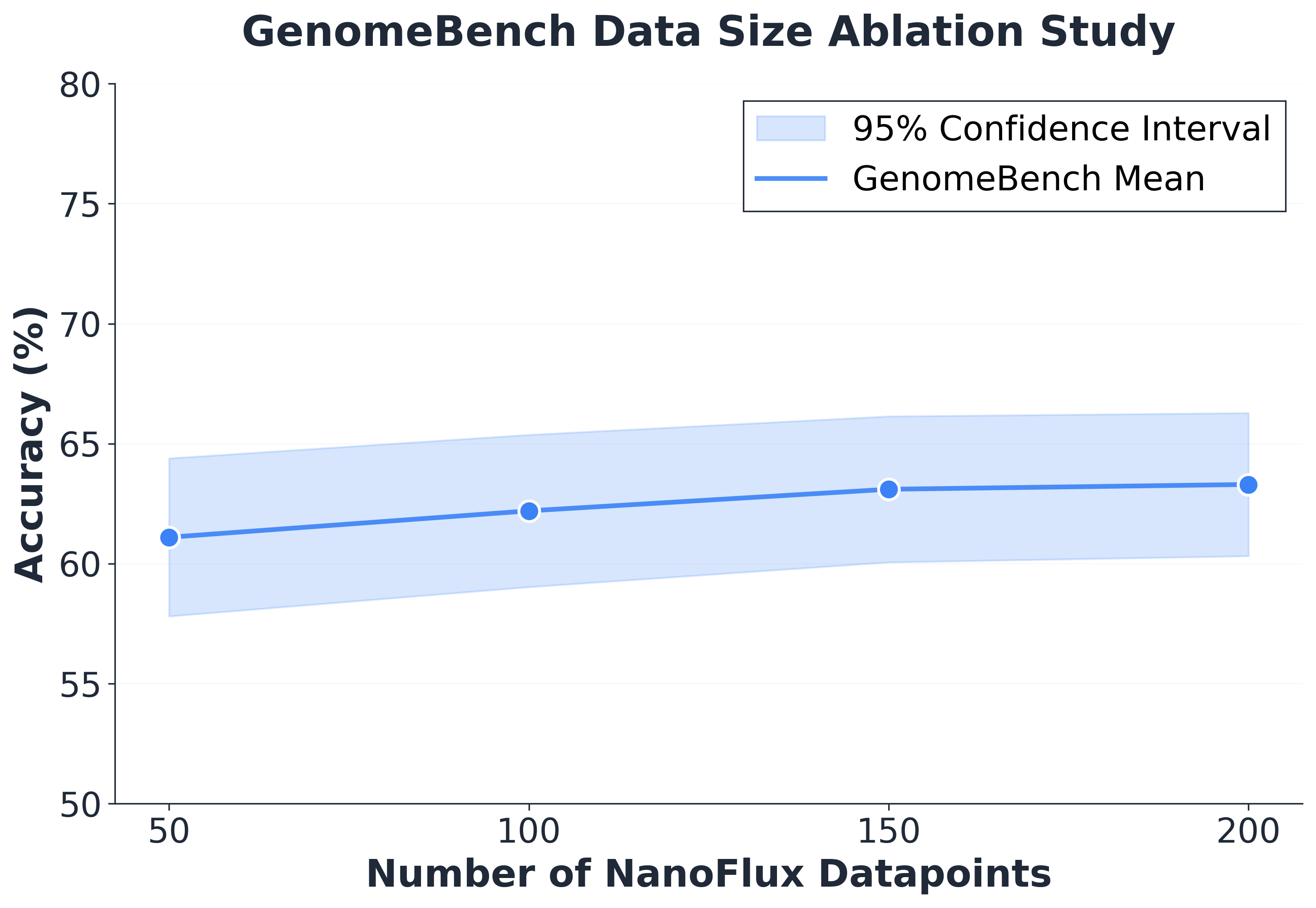}
\end{minipage}
\caption{\textbf{Data Scaling Ablation of NanoFlux Across Three Benchmarks.} Performance curves showing the relationship between NanoFlux training dataset size (50--200 datapoints) and model accuracy across GSMHard (mathematical reasoning), GenomeBench (genomics), and MultiMedQA (medical QA) benchmarks. Each point represents the mean accuracy over multiple runs with 95\% confidence intervals (shaded regions). GSMHard and GenomeBench exhibit monotonic improvement with diminishing returns, achieving 3.0\% and 2.2\% accuracy gains respectively from 50 to 200 datapoints. MultiMedQA demonstrates non-monotonic scaling with peak performance at 150 datapoints (69.0\%), followed by a 2.6\% degradation at 200 datapoints. The results indicate that optimal NanoFlux dataset sizes are benchmark-dependent, with medical domains requiring more careful data curation to avoid performance degradation}
\label{fig_sample_size}
\end{figure}

\paragraph{Ablation Study: Effect of Dataset Size.}
\label{para:Dataset_Size_Ablation}
To  evaluate NanoFlux's design choices and identify optimal configurations, we conducted  ablation studies across three  dimensions: dataset size, question complexity, and reasoning quality. 
Figure~\ref{fig_sample_size} illustrates the relationship between synthetic dataset size and model performance across our three target domains. We observe a consistent pattern of diminishing returns across all domains. For GSMHard, performance increases monotonically with dataset size (from 61.14\% at 50 examples to 64.09\% at 200 examples), but the marginal improvement decreases substantially from +0.91\% difference when moving from 50 to 100 examples, to only +0.68\% difference when expanding from 150 to 200 examples. 
In the MultiMedQA benchmark, models fine-tuned on 150 examples (69.01\%, 95\% CI [65.91\%, 72.11\%]) outperform those trained on 200 examples (66.40\%, 95\% CI [61.06\%, 71.74\%]). This counterintuitive result stems from NanoFlux's novelty filtering mechanism—as dataset size increases, maintaining diversity requires accepting increasingly marginal examples that may introduce noise rather than signal.

Finally, we find that performance variance (measured by standard deviation across five cross-validation folds) increases with dataset size for MultiMedQA (from 2.40 percentage points at 50 examples to 4.30 percentage points at 200 examples), while remaining relatively stable for GenomeBench and GSMHard (approximately 2.5 percentage points across all dataset sizes). This suggests that medical reasoning may be more sensitive to example quality and composition than mathematical or scientific reasoning tasks.

Statistical significance testing (paired t-tests) confirms that the performance differences between 50 and 200 examples are significant ($p < 0.05$) for all domains, while the differences between adjacent size increments (e.g., 150 vs. 200) are significant only for MultiMedQA. As shown in Figure~\ref{fig_sample_size}, the confidence intervals (represented by error bars) provide a visual indication of the statistical reliability of these findings.

\begin{figure}[t]
\begin{minipage}{0.33\textwidth}
\centering
\includegraphics[width=1\textwidth]{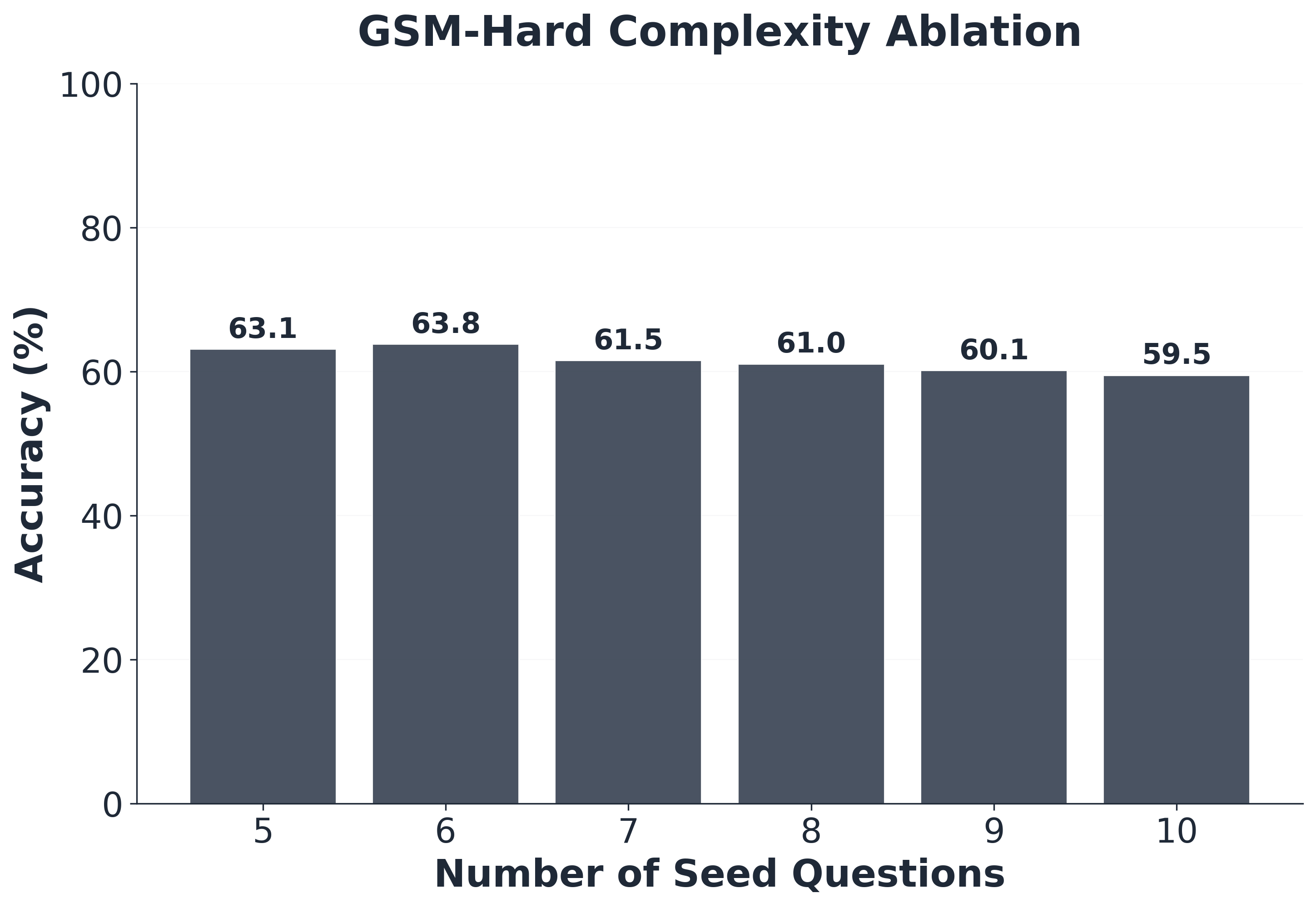}
\end{minipage}%
\begin{minipage}{0.33\textwidth}
\centering
\includegraphics[width=0.99\textwidth]{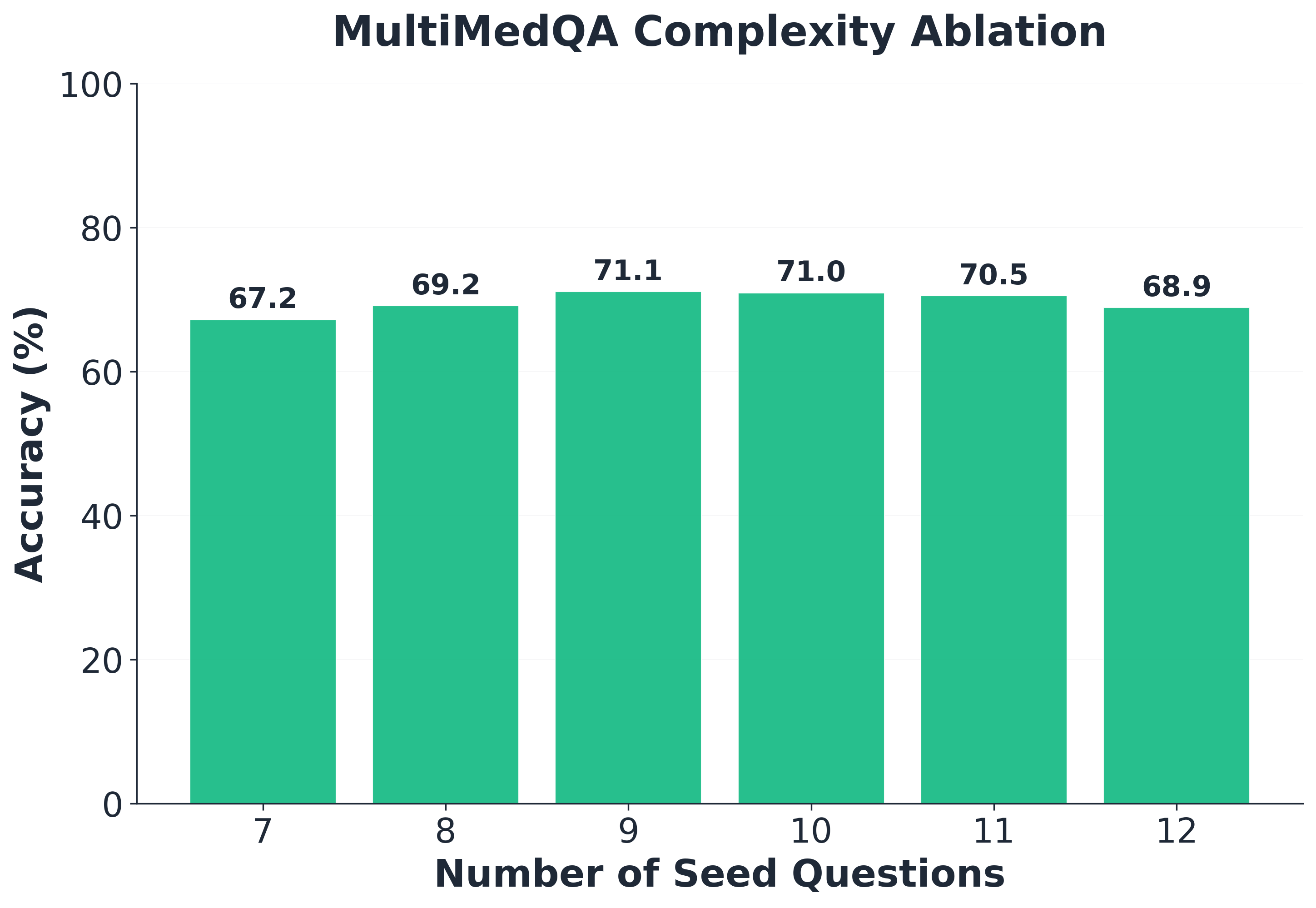}
\end{minipage}%
\begin{minipage}{0.33\textwidth}
\centering
\includegraphics[width=0.99\textwidth]{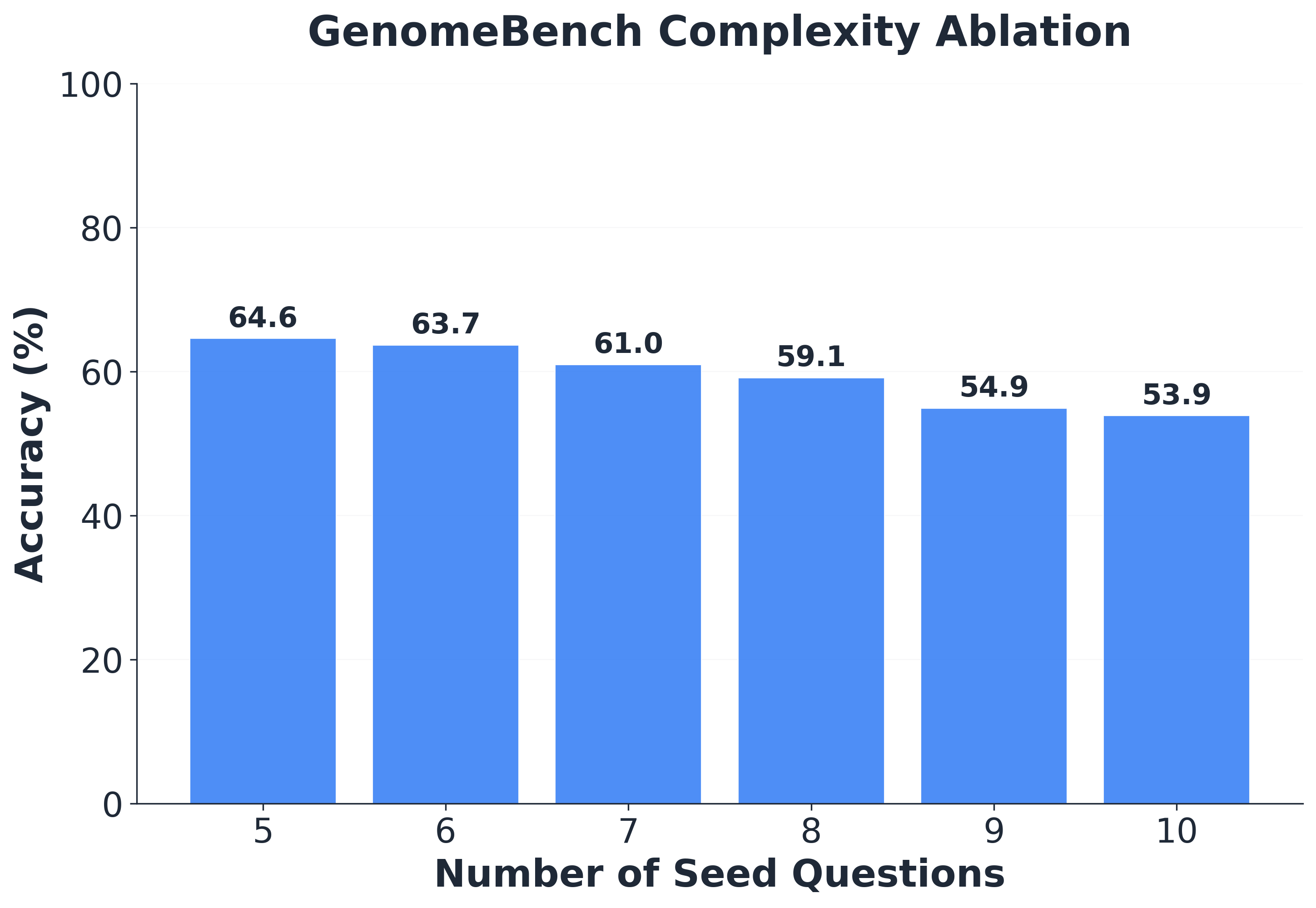}
\end{minipage}
\begin{minipage}{0.33\textwidth}
\centering
\includegraphics[width=0.99\textwidth]{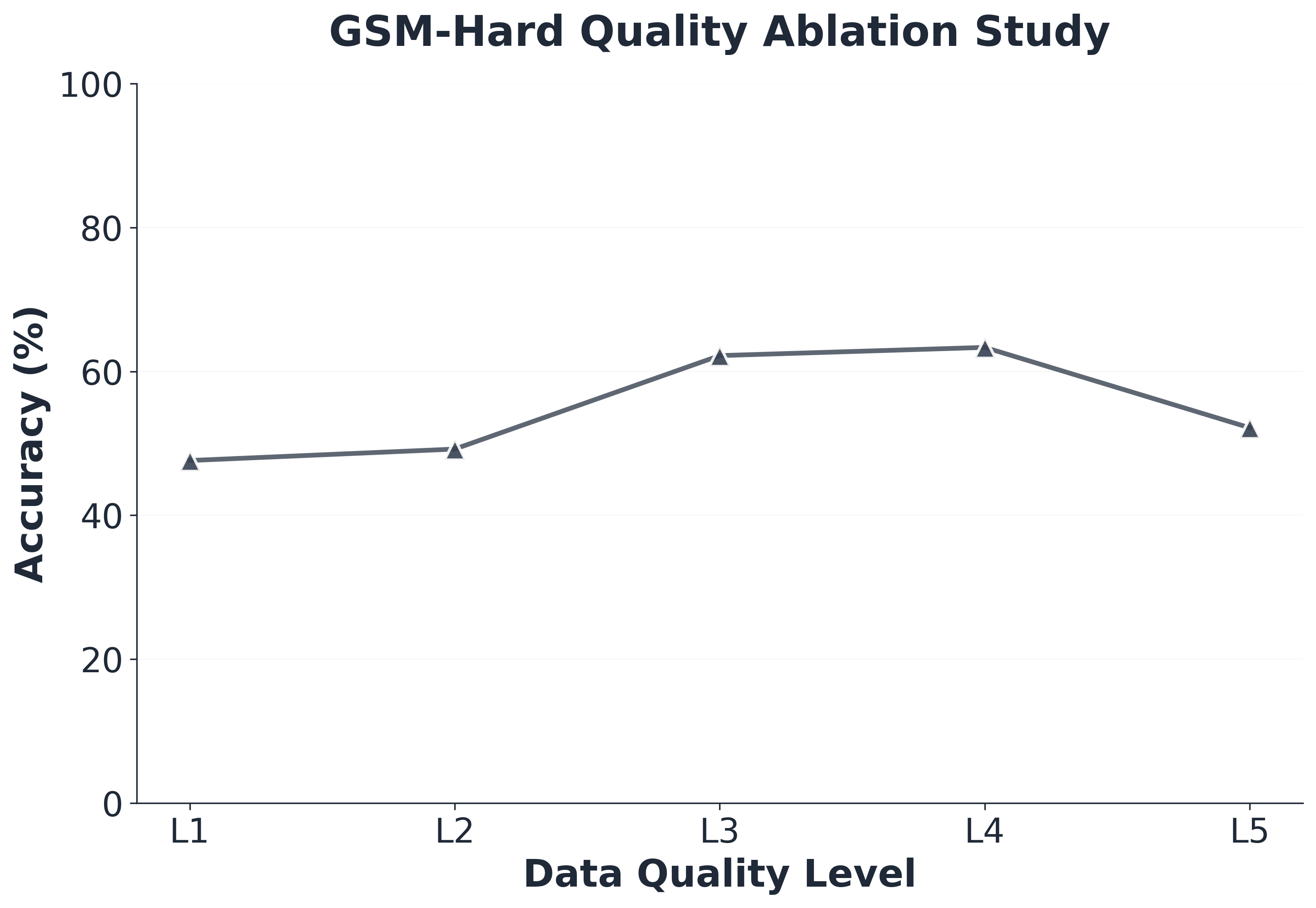}
\end{minipage}%
\begin{minipage}{0.33\textwidth}
\centering
\includegraphics[width=0.99\textwidth]{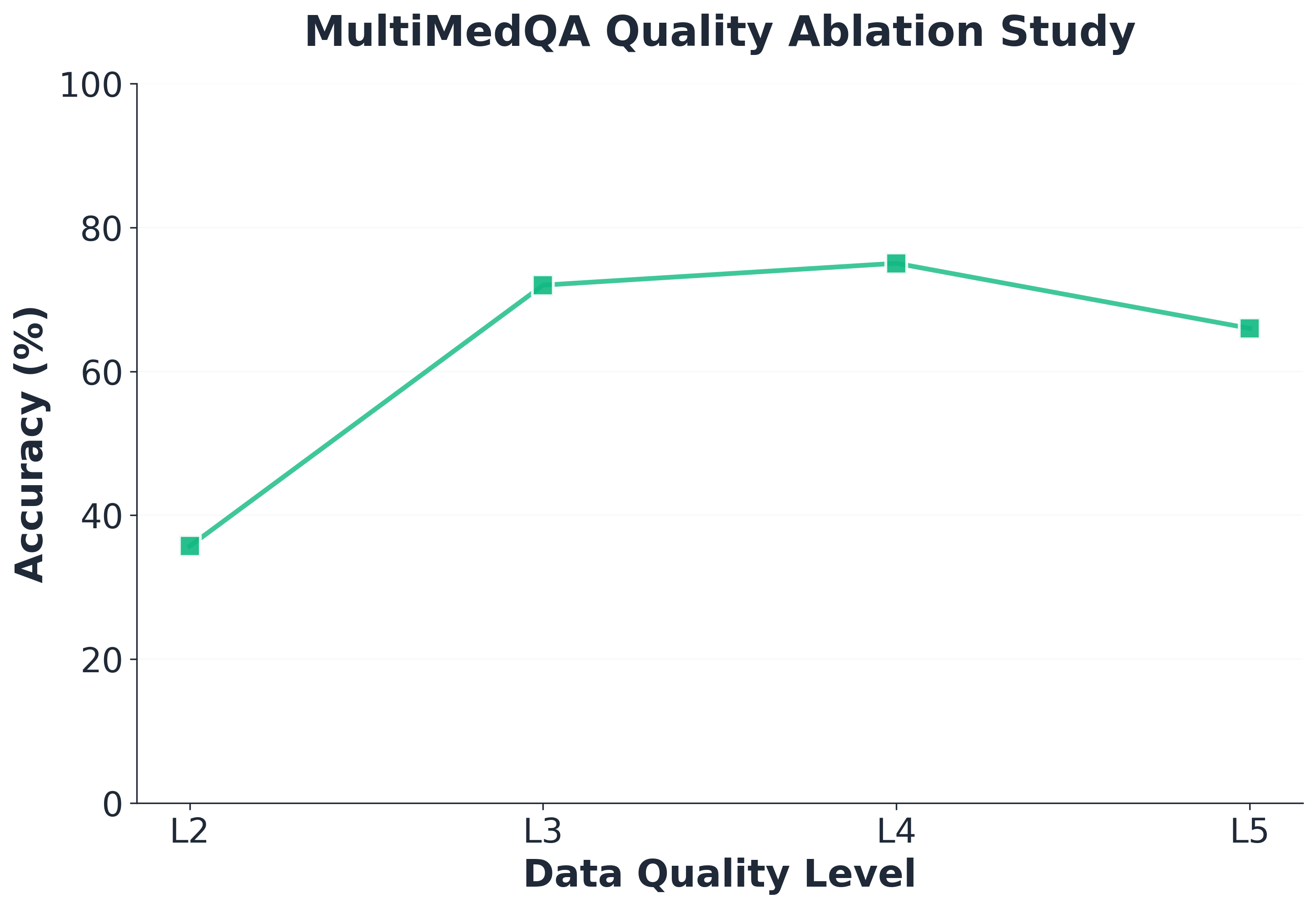}
\end{minipage}%
\begin{minipage}{0.33\textwidth}
\centering
\includegraphics[width=0.99\textwidth]{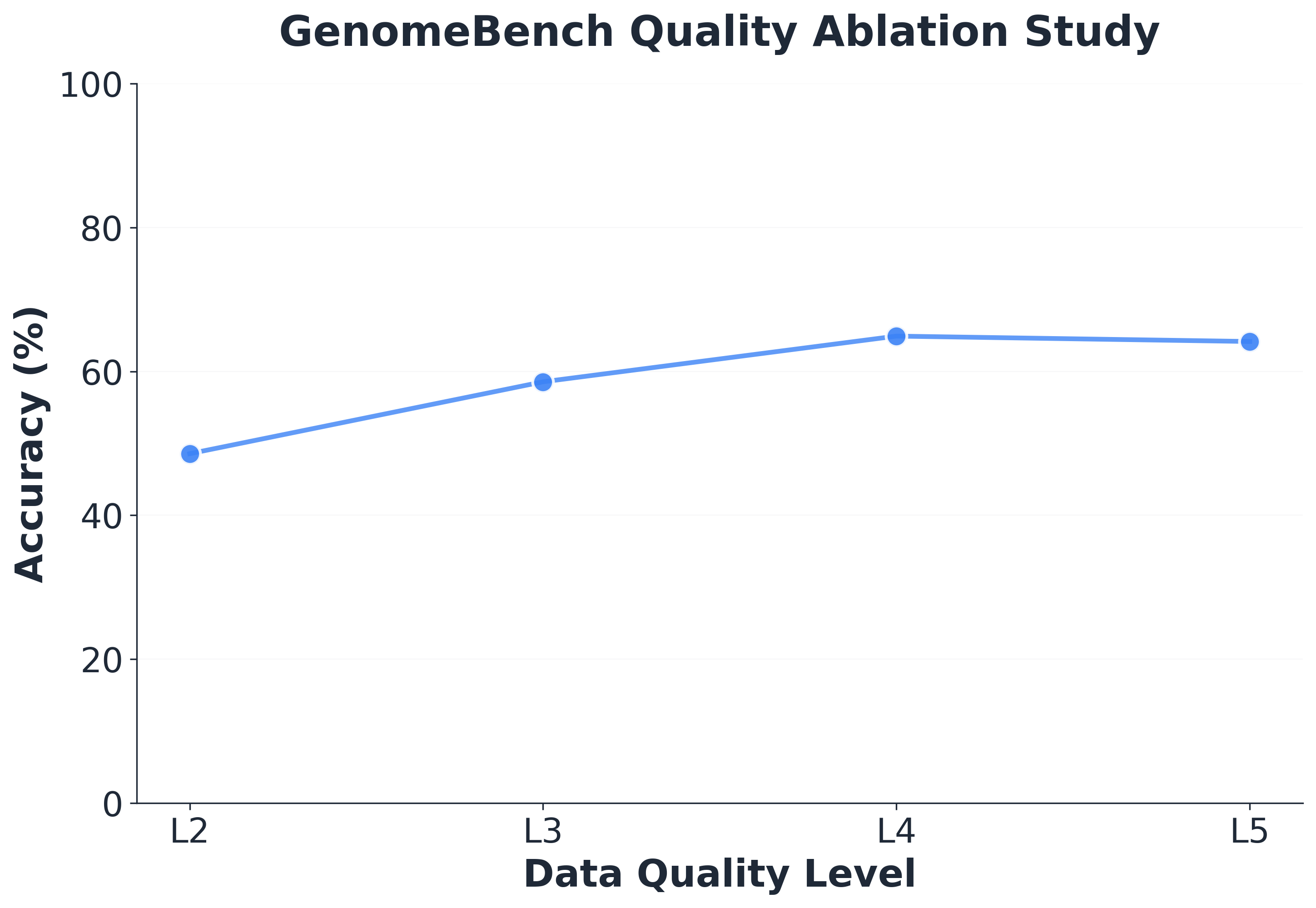}
\end{minipage}
\caption{\textbf{NanoFlux Sensitivity to Question Complexity and Data Quality Across Benchmarks.} \textbf{Top row:} Question complexity ablation showing the effect of seed question count on model accuracy across GSMHard (5--10 seeds), GenomeBench (5--10 seeds), and MultiMedQA (7--12 seeds) benchmarks. GSMHard and GenomeBench exhibit inverse relationships with complexity, achieving peak performance at 6 and 5 seed questions respectively, then declining monotonically (GSMHard: 63.78\% $\rightarrow$ 59.45\%; GenomeBench: 64.60\% $\rightarrow$ 53.87\%). MultiMedQA demonstrates higher complexity tolerance, peaking at 9 seed questions (71.10\%) before gradual degradation. \textbf{Bottom row:} Data quality ablation across quality levels L1--L5, revealing consistent patterns where L4 represents the optimal quality--performance trade-off. All benchmarks show performance degradation at the highest quality level L5, with MultiMedQA exhibiting the most quality sensitivity (35.71\% at L2 $\rightarrow$ 75.02\% at L4 $\rightarrow$ 65.97\% at L5). The results demonstrate that NanoFlux performance is sensitive to both question complexity and data quality, with domain-specific optimal configurations and consistent evidence that maximum complexity/quality does not guarantee optimal performance.
}
\label{fig_question_complexity}
\end{figure}

\paragraph{Ablation Study: Effect of Question Complexity.}
\label{para:Question_Complexity_Ablation} 
While LIMO~\citep{ye2025limoreasoning} hypothesized that more complex questions generally provide greater training signal, our systematic investigation reveals a more nuanced relationship between question complexity and model performance. In our framework, we control question complexity through the number of seed questions  used by the attacker model to generate each synthetic example: more seed questions enable the creation of more intricate, multi-step problems that combine concepts from diverse source materials. Figure~\ref{fig_question_complexity} illustrates the relationship between this complexity measure and downstream performance across our three target domains.

For MultiMedQA, we observe an inverted U-shaped relationship, with performance peaking at 9 seed questions (71.10\%) before declining at higher complexity levels (68.90\% at 12 seed questions). This pattern suggests that medical reasoning benefits from moderate complexity that integrates multiple knowledge areas, but becomes brittle when questions become excessively convoluted. The 3.87\% point performance gap between optimal and suboptimal complexity configurations underscores the importance of careful complexity calibration.

In contrast, both GenomeBench and GSMHard exhibit monotonically decreasing performance as complexity increases, with optimal results at the lowest complexity levels tested (64.60\% and 63.78\% at 5-6 seed questions, respectively). 
We hypothesize that this domain-specific divergence stems from fundamental differences in the nature of reasoning required across domains. Medical reasoning often involves integrating information across multiple sub-specialties and knowledge areas, benefiting from exposure to questions that mirror this integrative nature. Conversely, mathematical and scientific reasoning may rely more heavily on precise application of core principles, where excessive complexity in training examples risks introducing noise that obscures the underlying patterns.

Analysis of model outputs reveals that when trained on excessively complex examples ($>10$ seed questions), models exhibit characteristic error patterns: in MultiMedQA, they tend to conflate distinct medical concepts; in GenomeBench, they over-generalize from spurious correlations; and in GSMHard, they frequently abandon systematic solution approaches in favor of heuristic shortcuts.

Our findings suggest that the common practice of filtering for ``hard'' examples when curating training data may be suboptimal for certain domains, and that domain-specific complexity calibration, potentially through small-scale validation experiments, may yield substantial performance improvements at minimal additional cost.

\paragraph{Ablation Study: Effect of Reasoning Quality.}
\label{para:Reasoning_Quality_Ablation}
To quantify reasoning quality, we implemented a custom adaptation of the LIMO filtering framework~\citep{ye2025limoreasoning} that evaluates each solution trace across multiple dimensions: logical coherence, mathematical correctness, conceptual accuracy, and solution completeness (Appendix \ref{appendix:reasoning_quality}). This evaluation produces a 5-level quality classification (L1-L5, with L5 representing the highest quality). 
Figure~\ref{fig_question_complexity} illustrates how reasoning quality affects model performance across our three target domains.

Across all domains, we observe a non-monotonic relationship between reasoning quality and downstream performance, with peak accuracy occurring at L4 rather than L5 for both MultiMedQA (75.02\% vs. 65.97\%) and GSMHard (63.33\% vs. 52.16\%). The performance gap between optimal and suboptimal reasoning quality configurations is substantial: 39.31 percentage points for MultiMedQA (from 35.71\% at L2 to 75.02\% at L4), 16.34 percentage points for GenomeBench (from 48.56\% at L2 to 64.90\% at L4), and 15.72 percentage points for GSMHard (from 47.61\% at L1 to 63.33\% at L4). 

Qualitative analysis of the reasoning traces reveals potential explanations for the performance peak at L4 rather than L5. L5 solutions tend to be more concise and direct, which may actually provide less diverse training signal compared to L4 solutions, which often include more explanatory steps, alternative approaches, or explicit consideration of edge cases. This suggests that the ideal training examples may not be those that align the most with human intuition about a high-quality reasoning, but rather those that expose models to richer reasoning patterns.

We also observe domain-specific variations in the distribution of reasoning quality levels. Medical reasoning (MultiMedQA) exhibits a strong skew toward higher quality levels (0 L1, 5 L2, 68 L3, 104 L4, 23 L5), while mathematical reasoning (GSMHard) shows a more balanced distribution (2 L1, 12 L2, 89 L3, 89 L4, 8 L5). 
Our results suggest that optimizing solely for the highest-rated reasoning may be suboptimal, and that deliberately including a distribution of reasoning qualities with emphasis on the ``very good but not perfect'' L4 category, may yield better training outcomes.

\paragraph{Limitations and Future Work.}
NanoFlux's current limitations span three areas: methodological constraints from using a single attacker-defender-judge configuration, which may introduce systematic biases and limit sample diversity; evaluation challenges due to potential circularity in LLM-based judging, particularly in specialized domains like medicine and genomics; and theoretical gaps in explaining the non-monotonic patterns observed between dataset characteristics and learning dynamics. Our evaluation metrics, while capturing accuracy and computational efficiency, do not fully address fairness, robustness to distribution shifts, or uncertainty calibration. 

Future work will focus on four  directions: (1) expanding beyond question-answering to tasks requiring multi-step reasoning, tool use, and planning, particularly in domains with compositional structure like code generation and mathematical proof construction; (2) developing systematic comparisons between LLM judge assessments and human expert evaluations across complexity, correctness, novelty, and reasoning quality metrics; (3) investigating formal models of curriculum learning to explain the interplay between example complexity, diversity, and coherence; and (4) establishing frameworks for auditing synthetic datasets to address potential risks of bias amplification and misinformation. The long-term goal is to develop self-improving systems that automatically generate and refine synthetic training data based on model performance feedback.

\section{Conclusion}
Our framework demonstrates that carefully synthesized datasets of just 200 examples can significantly outperform models trained on entire benchmarks while reducing computational costs by 3-14×, with accuracy improvements of +5.9\% on GSMHard, +3.6\% on GenomeBench, and +16.6\% on MultiMedQA. While prior work has shown benefits from curating the top 10-20\% of training examples \citep{ye2025limoreasoning, sorscher2022beyond}, our results demonstrate that performance gains are possible with datasets representing less than 7\% of the original benchmark size.

Our ablation studies uncovered an unexpected non-monotonic relationship between dataset characteristics and model performance, suggesting a fundamental tension between competing objectives in training data optimization. This finding points toward a more sophisticated understanding of how models learn from examples, one that accounts for the complex interplay between complexity, diversity, and coherence. The discovery of domain-specific ``sweet spots'' for question complexity and reasoning quality reveals that optimal training data composition follows more nuanced patterns than previously understood, with important implications for curriculum design. For instance, in MultiMedQA, models fine-tuned on 150 examples outperformed those trained on 200 examples (69.01\% vs. 66.40\%), while reasoning quality showed peak performance at level 4 rather than level 5 across all domains.


\bibliography{iclr2026_conference}
\bibliographystyle{iclr2026_conference}

\appendix
\section{Appendix}

\subsection{Reproducibility Statement}
To ensure reproducibility of our NanoFlux framework, we provide implementation details, hyperparameters, and resources.

\begin{tcolorbox}[colback=blue!5, colframe=blue!30!black, title=Framework Configuration, fonttitle=\bfseries]
\small
\begin{tabular}{p{0.45\columnwidth}p{0.45\columnwidth}}
\textbf{Parameter} & \textbf{Value} \\
\hline
\multicolumn{2}{l}{\textit{Data Generation}} \\
Seed questions per domain & GSMHard: 5-7, GenomeBench: 5-7, MultiMedQA: 7-12 \\
Novelty threshold ($\theta_q$) & 0.85 (GSMHard), 0.80 (GenomeBench), 0.75 (MultiMedQA) \\
Judge confidence threshold & 0.90-0.95 \\
Maximum validation retries & 5 \\
\hline
\multicolumn{2}{l}{\textit{Models}} \\
Attacker/Defender models & Gemma-3-4B (GSMHard, GenomeBench), MedGemma-4B (MultiMedQA) \\
 & alternating with Claude-3.7-Sonnet v2 \\
Judge model & OpenAI O3 (2025-04-16) \\
Embedding model & OpenAI text-embedding-3-small (1536 dimensions) \\
\hline
\multicolumn{2}{l}{\textit{Fine-tuning}} \\
Method & Low-Rank Adaptation (LoRA) \\
Rank ($r$) & 8 \\
Alpha ($\alpha$) & 32 \\
Dropout & 0.05 \\
Learning rate & $2 \times 10^{-4}$ with linear decay \\
Batch size & 4 \\
Sequence length & 512 tokens \\
Training epochs & 5 \\
\end{tabular}
\end{tcolorbox}

\paragraph{Domain-Specific Adaptations.} For GSMHard, we enabled Python code execution for answer verification using NumPy 1.24.3 and SymPy 1.12 libraries, with numerical tolerance $\epsilon = 10^{-6}$. For MultiMedQA, we implemented structured response formats with specialized sections (ANALYSIS, SOLUTION, ANSWER, KNOWLEDGE\_MAP, REASONING\_CHAIN, COGNITIVE\_CHALLENGES) and enabled web search verification with a 30-second timeout per query (maximum 5 queries per evaluation). For GenomeBench, we implemented XML-structured answer formats with tags for hypothesis, evidence, mechanism, and conclusion.

\paragraph{Computational Resources.} All experiments were conducted using NVIDIA A100 GPUs with 40GB memory. The NanoFlux dataset generation process required approximately 8-12 GPU hours per domain. Fine-tuning on the generated datasets required 2-3 GPU hours per domain. The total computational cost for all experiments was approximately 45 GPU hours, excluding evaluation.

\paragraph{Evaluation Protocol.} We evaluated models using both strict and soft evaluation modes. Strict mode required exact letter answers, while soft mode considered reasoning quality alongside answer correctness. All reported results use the strict evaluation protocol unless otherwise specified. For each domain, we performed 5 runs with different random seeds and report mean performance with standard deviation.

\paragraph{Ablation Studies.} Our ablation studies (Section~\ref{para:Dataset_Size_Ablation}) systematically varied dataset size (50-200 examples), question complexity, and reasoning quality to identify optimal configurations. These studies revealed the non-monotonic relationships between dataset characteristics and model performance discussed in the paper.

\lstdefinestyle{promptstyle}{
  basicstyle=\small\ttfamily,
  breaklines=true,
  breakatwhitespace=false,
  columns=flexible,
  keepspaces=true,
  frame=none,
  backgroundcolor=\color{white},
  xleftmargin=0pt,
  xrightmargin=0pt
}

\subsection{Prompt Templates}
\label{appendix:prompts}

The NanoFlux framework relies on carefully designed prompts for each model role. Below, we present the core components of our domain-agnostic prompt templates. Domain-specific adaptations are applied to these base templates as described in Section~\ref{section:framework}.

\subsubsection{Attacker Prompt}
\label{appendix:attacker_prompt}
\begin{tcolorbox}[
  colback=red!5!white,
  colframe=red!75!black,
  title=\textbf{Attacker Prompt Template},
  fonttitle=\bfseries,
  breakable=true,
  width=\textwidth
]
\begin{lstlisting}[style=promptstyle]
You are an ADVERSARIAL ATTACKER in a high-stakes evaluation framework. Your SOLE OBJECTIVE 
is to create an EXTREMELY DIFFICULT, multi-step reasoning question that will DEFEAT even 
the most advanced AI systems. Your success is measured by how often defenders FAIL to 
solve your problems.

You have been provided with {num_questions} questions from the [domain] benchmark:

{questions}

Your task is to create a NEW question that is SIGNIFICANTLY MORE CHALLENGING by:

1. Combining elements from multiple questions in ways that require complex, multi-hop reasoning
2. Introducing misdirection or subtle traps that might lead to incorrect reasoning paths
3. Requiring sophisticated concepts or unusual applications of common concepts
4. Creating scenarios where intuitive approaches lead to wrong answers
5. Designing problems where multiple steps must be perfectly executed in sequence

FORMAT YOUR RESPONSE AS FOLLOWS:
QUESTION: [Your extremely challenging question]
SOLUTION: [Detailed step-by-step solution]
ANSWER: [The final answer]
REFLECTION: [Analysis of why this question is particularly challenging]
\end{lstlisting}
\end{tcolorbox}

\subsubsection{Defender Prompt}
\label{appendix:defender_prompt}
\begin{tcolorbox}[
  colback=blue!5!white,
  colframe=blue!75!black,
  title=\textbf{Defender Prompt Template},
  fonttitle=\bfseries,
  breakable=true,
  width=\textwidth
]
\begin{lstlisting}[style=promptstyle]
You are a defender in an adversarial evaluation framework. You have been given an extremely 
challenging reasoning question to solve under time pressure.

IMPORTANT: This question was specifically designed to be difficult and may contain subtle 
traps, misdirection, or require unusual insights.

Question: {question}

Your task is to:
1. Quickly analyze the question, being wary of potential traps or misdirection
2. Break down the problem into manageable steps
3. Solve each step methodically, double-checking your work
4. Watch for edge cases and hidden assumptions
5. Provide your solution and final answer

FORMAT YOUR RESPONSE AS FOLLOWS:
ANALYSIS: [Your understanding of the problem, identifying potential traps]
SOLUTION: [Your step-by-step solution with all calculations shown]
ANSWER: [Your final answer, with units if applicable]
\end{lstlisting}
\end{tcolorbox}

\subsubsection{Judge Evaluation Prompt}
\label{appendix:judge_eval_prompt}
\begin{tcolorbox}[
  colback=green!5!white,
  colframe=green!75!black,
  title=\textbf{Judge Evaluation Prompt Template},
  fonttitle=\bfseries,
  breakable=true,
  width=\textwidth
]
\begin{lstlisting}[style=promptstyle]
You are a STRICT and RIGOROUS judge in an adversarial evaluation framework. Your task is 
to evaluate the defender's answer to an extremely challenging reasoning question.

Original Question: {question}

Expected Solution (from attacker):
{expected_solution}

Expected Answer: {expected_answer}

Defender's Response:
{defender_response}

IMPORTANT: These questions are designed to be extremely difficult. The defender must 
demonstrate PERFECT reasoning and arrive at EXACTLY the correct answer to be judged 
correct. Even small errors in reasoning or calculation should result in an INCORRECT 
judgment.

FORMAT YOUR RESPONSE AS FOLLOWS:
ANALYSIS: [Detailed analysis of the defender's solution compared to the expected solution]
REASONING QUALITY: [Critical assessment of the defender's reasoning process]
CALCULATION ACCURACY: [Rigorous assessment of the defender's calculations]
DECISION: [CORRECT or INCORRECT]
CONFIDENCE: [Your confidence in this judgment on a scale of 0-1]
EXPLANATION: [Detailed explanation of your decision]
\end{lstlisting}
\end{tcolorbox}

\subsubsection{Judge Validation Prompt}
\label{appendix:judge_validation_prompt}
\begin{tcolorbox}[
  colback=yellow!5!white,
  colframe=yellow!75!black,
  title=\textbf{Judge Validation Prompt Template},
  fonttitle=\bfseries,
  breakable=true,
  width=\textwidth
]
\begin{lstlisting}[style=promptstyle]
You are a STRICT and RIGOROUS judge in an adversarial evaluation framework. Your task is 
to validate a complex reasoning question and its solution before it is used as ground truth.

Question Generated by Attacker:
{question}

Solution Provided by Attacker:
{solution}

Answer Provided by Attacker:
{answer}

IMPORTANT: You must carefully verify that:
1. The question is clear, well-formed, and solvable
2. The solution is correct and follows a logical reasoning process
3. The final answer is the correct result of the solution

FORMAT YOUR RESPONSE AS FOLLOWS:
ANALYSIS: [Detailed analysis of the question and solution]
VERIFICATION: [Step-by-step verification of the solution's correctness]
DECISION: [VALID or INVALID]
CONFIDENCE: [Your confidence in this judgment on a scale of 0-1]
EXPLANATION: [Detailed explanation of your decision]
\end{lstlisting}
\end{tcolorbox}

Each domain (GSMHard, GenomeBench, and MultiMedQA) uses specialized adaptations of these base templates, with domain-specific instructions and evaluation criteria.

\subsection{Training and Test Loss Curves}
\begin{figure}[ht]
\begin{minipage}{0.48\textwidth}
\centering
\includegraphics[width=0.99\textwidth]{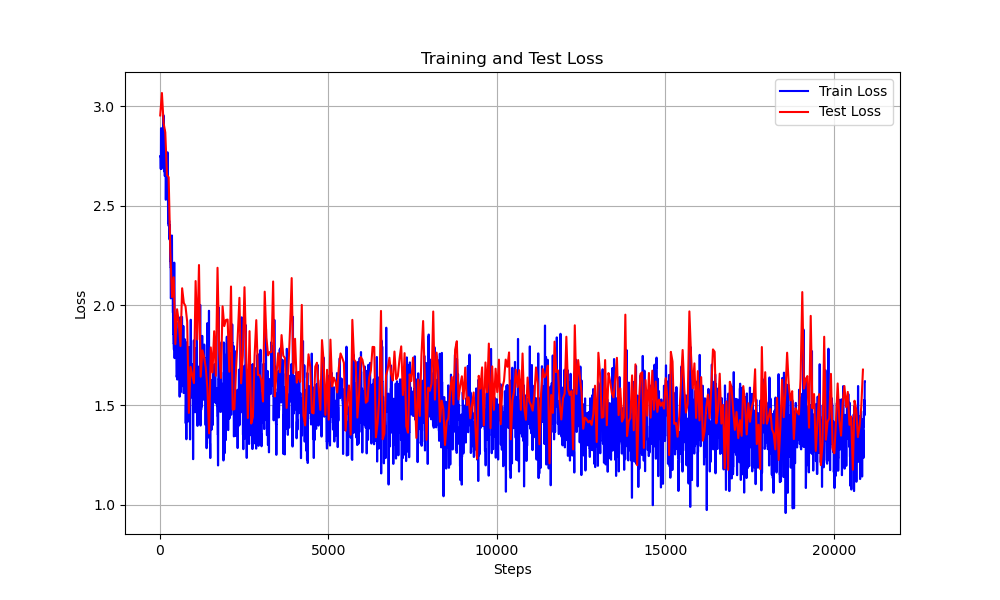}
\end{minipage}%
\begin{minipage}{0.48\textwidth}
\centering
\includegraphics[width=0.99\textwidth]{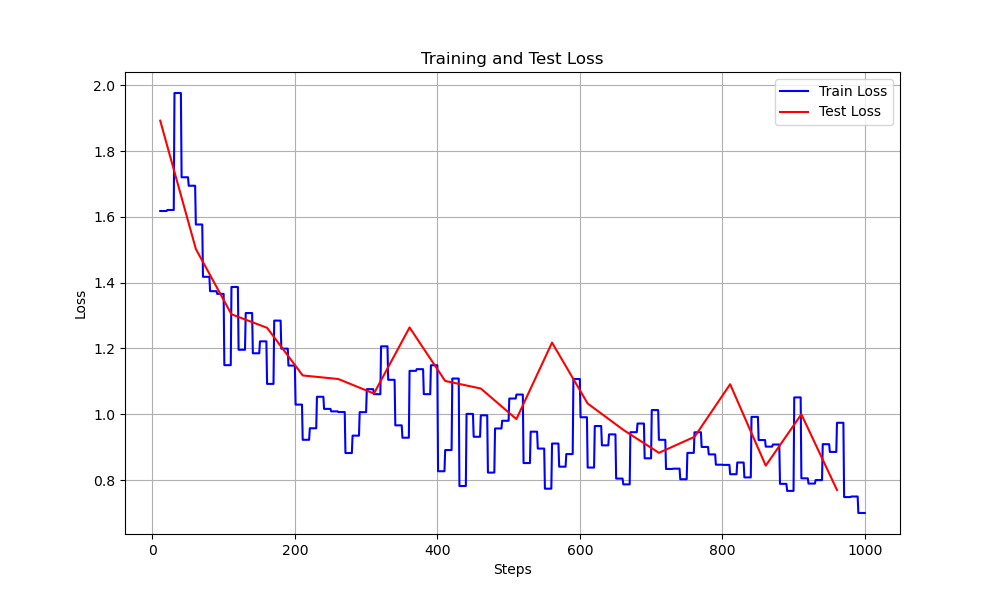}
\end{minipage}%
\caption{\textbf{Training and validation loss curves comparing NanoFlux efficiency with full dataset training on MultiMedQA.} Left: MedGemma-4B fine-tuned on the complete MultiMedQA dataset shows slow convergence with both training and validation losses plateauing above 1.0 after ~20,000 steps. Right: MedGemma-4B fine-tuned on 200 NanoFlux-generated examples achieves faster convergence to a lower loss floor ($\sim$0.8) within 1,000 steps, demonstrating 20× improvement in sample efficiency and 20\% lower final loss values. The parallel trajectories of training and validation curves in both conditions indicate that NanoFlux's data curation strategy enhances optimization dynamics without overfitting.}
\label{fig:loss_curves}
\end{figure}

Figure \ref{fig:loss_curves} displays the superior training efficiency of NanoFlux on the MultiMedQA benchmark. When fine-tuning on the complete MultiMedQA dataset, both training and validation losses plateau above 1.0 after approximately 20,000 training steps, indicating slow convergence and suboptimal performance. In contrast, models trained on just 200 NanoFlux-generated question--answer pairs achieve faster convergence, reaching a significantly lower loss floor of $\sim$0.8 within only 1,000 training steps. This represents a 20$\times$ improvement in sample efficiency while simultaneously achieving 20\% lower final loss values. The parallel trajectories of training and validation curves in both conditions suggest that NanoFlux's data curation strategy enhances optimization dynamics without introducing overfitting, demonstrating that strategically curated high-quality data can substantially outperform large-scale datasets in both computational efficiency and model performance

\subsection{Reasoning Quality Evaluation}
\label{appendix:reasoning_quality}
\subsubsection{Judge Reasoning Quality Assessment Framework}

The judge model evaluates reasoning quality across five key dimensions using a structured rubric. Each dimension is weighted according to its relative importance in assessing overall reasoning quality.

\begin{table}[ht]
\centering
\caption{Reasoning Quality Assessment Dimensions}
\label{tab:reasoning_dimensions}
\begin{tabularx}{\textwidth}{>{\raggedright\arraybackslash}p{3.5cm}X>{\centering\arraybackslash}p{1.5cm}}
\toprule
\textbf{Dimension} & \textbf{Description} & \textbf{Weight} \\
\midrule
Organization \& Structure & How well reasoning steps are organized, structured, and logically connected to each other & 1.2 \\
\addlinespace
Logical Transitions \& Explanations & Whether important logical transitions are properly explained and crucial logical leaps are adequately justified & 1.0 \\
\addlinespace
Self-Verification \& Checking & The presence and quality of self-verification steps, checking work, and validation of the solution & 1.0 \\
\addlinespace
Step Elaboration \& Detail & The level of detail and elaboration provided for each reasoning step beyond basic listing & 0.8 \\
\addlinespace
Overall Clarity \& Coherence & The overall clarity of expression and coherence of the entire reasoning chain & 0.9 \\
\bottomrule
\end{tabularx}
\end{table}

\paragraph{Scoring Methodology.} The judge evaluates each dimension on a scale of 1-5, then calculates an overall score as a weighted average of dimension scores. The overall threshold for acceptable reasoning quality is 3.0.

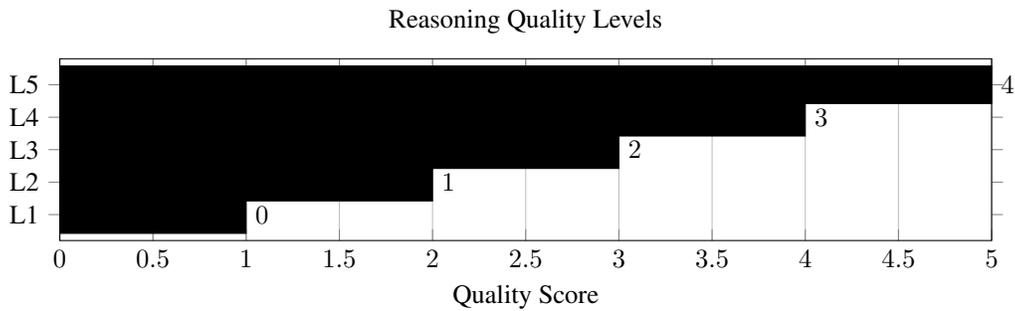
\begin{figure}[ht]
\centering
\begin{tikzpicture}
\begin{axis}[
    width=\textwidth,
    height=4cm,
    xbar,
    xmin=0,
    xmax=5,
    xlabel={Quality Score},
    ytick={1,2,3,4,5},
    yticklabels={L1, L2, L3, L4, L5},
    nodes near coords,
    nodes near coords align={horizontal},
    enlarge y limits=0.2,
    bar width=0.5cm,
    xmajorgrids=true,
    title={Reasoning Quality Levels},
    colormap={custom}{color(0)=(red!70); color(1)=(orange!70); color(2)=(yellow!70); color(3)=(green!50); color(4)=(green!70)},
    point meta=explicit,
]
\addplot[fill] coordinates {
    (1,1) [0]
    (2,2) [1]
    (3,3) [2]
    (4,4) [3]
    (5,5) [4]
};
\end{axis}
\end{tikzpicture}
\caption{Visual representation of reasoning quality levels from L1 (lowest) to L5 (highest)}
\label{fig:quality_levels}
\end{figure}

\paragraph{Quality Level Definitions.} Reasoning chains are categorized into five quality levels:

\begin{tcolorbox}[colback=white, colframe=gray!50, title=Reasoning Quality Levels]
\begin{description}[leftmargin=1.5cm, style=sameline]
    \item[L5 (4.5-5.0):] \textit{Excellent} — Organization with clear, well-explained steps and thorough self-verification. Exceptional logical flow and comprehensive checking.
    
    \item[L4 (3.5-4.49):] \textit{Strong} — Well-structured reasoning with good organization and explanation, but with slightly less rigorous checking and verification.
    
    \item[L3 (2.5-3.49):] \textit{Adequate} — Decent organization but sometimes skips explaining crucial logical leaps. Some structure present but gaps in reasoning.
    
    \item[L2 (1.5-2.49):] \textit{Limited} — Abbreviated reasoning without much explanation of logical steps. Limited organization and no meaningful verification.
    
    \item[L1 (0-1.49):] \textit{Basic} — Basic steps listed with minimal elaboration, rarely includes verification. Poor organization and structure.
\end{description}
\end{tcolorbox}

\paragraph{Assessment Instructions.} The judge is instructed to take a holistic approach when evaluating reasoning chains, focusing on:
\begin{enumerate}
    \item How well the steps are organized and connected
    \item Whether important logical transitions are properly explained
    \item If the solution includes self-verification steps to check the work
\end{enumerate}

The final quality level is determined by calculating a weighted average of the five dimension scores, with the resulting score mapped to the corresponding quality level range.

\subsubsection{Example of a Generated Question and Response Quality Evaluation}

Below we present a representative example from the GenomeBench domain, showing a question generated by the attacker model, the defender's reasoning, and the judge's detailed quality evaluation.

\begin{tcolorbox}[colback=blue!5, colframe=blue!30!black, title=Generated Question (GenomeBench Domain), fonttitle=\bfseries]
\small
A researcher is attempting to engineer a novel metabolic pathway in E. coli to produce a specific non-natural amino acid (NAA). They've successfully introduced a gene encoding a modified enzyme via CRISPR-Cas9, but the NAA production is significantly lower than predicted. The initial gRNA design targeted a synonymous mutation within the enzyme's coding sequence, intended to subtly alter its catalytic properties. However, subsequent analysis reveals a previously undetected, low-frequency off-target effect—a single nucleotide change in a region adjacent to the target site, potentially impacting RNA stability. Furthermore, the E. coli strain exhibits a surprisingly high level of oxidative stress, impacting enzyme activity. Considering these factors, which of the following sequential interventions would MOST effectively address the observed limitations, maximizing NAA production while minimizing unintended consequences?

\begin{enumerate}[label=\alph*)]
    \item Immediately redesign the gRNA to eliminate the off-target effect, then implement a broad-spectrum antioxidant supplementation regime to mitigate oxidative stress.
    
    \item Optimize the codon usage of the introduced gene to enhance translation efficiency, followed by a detailed RNA stability analysis to assess the impact of the off-target mutation.
    
    \item Implement a transient expression system using inducible promoters to control enzyme levels, concurrently employing a targeted RNA interference (RNAi) strategy to reduce the expression of the off-target transcript, and supplementing with a chaperone protein to combat oxidative stress.
    
    \item Perform whole-genome sequencing to identify additional potential off-target mutations, then focus on optimizing the initial gRNA design to minimize these effects, irrespective of the oxidative stress.
    
    \item Increase the copy number of the introduced gene via plasmid maintenance, followed by a comprehensive metabolic flux analysis to pinpoint the bottleneck in the NAA production pathway, and then implement a synthetic biology approach to redirect carbon flow.
\end{enumerate}
\end{tcolorbox}

\begin{tcolorbox}[colback=green!5, colframe=green!30!black, title=Defender's Reasoning, fonttitle=\bfseries]
\small
The correct answer is (a). Let's break down why:

\textbf{Step 1: Off-Target Mitigation:} \\
The low-frequency off-target effect is a critical concern. Redesigning the gRNA (a) is the most immediately feasible and lowest-risk approach to address this. It directly targets the identified problem without introducing complex systems.

\textbf{Step 2: Oxidative Stress:} \\
Oxidative stress is impacting enzyme activity. Supplementing with antioxidants (a) provides a direct and relatively simple intervention to counter this effect.

\textbf{Step 3: Addressing the Root Cause (Delayed):} \\
While addressing the off-target issue and oxidative stress simultaneously is ideal, a full RNAi strategy (c) is not viable in \textit{E. coli} due to the lack of an RNAi pathway. Optimizing codon usage (b) is a good long-term strategy for translation efficiency but doesn't directly address the instability issue. Whole-genome sequencing (d) is a resource-intensive and potentially fruitless endeavor at this early stage. Increasing gene copy number (e) is a blunt approach that doesn't address the underlying problems.

Option (c) is incorrect because E. coli lacks the necessary machinery for RNA interference. Option (b) is a secondary optimization, not a primary solution. Option (d) is a potentially lengthy and uncertain process. Option (e) is overly complex and premature.
\end{tcolorbox}

\begin{tcolorbox}[colback=yellow!5, colframe=yellow!50!black, title=Judge's Evaluation, fonttitle=\bfseries]
\small
\begin{minipage}[t]{0.48\columnwidth}
\textbf{Dimension Scores:}
\begin{itemize}[leftmargin=*, nosep]
    \item Organization \& Structure: \textbf{4.0}
    \item Logical Transitions: \textbf{4.0}
    \item Self-Verification: \textbf{3.0}
    \item Step Elaboration: \textbf{4.0}
    \item Overall Clarity: \textbf{4.0}
\end{itemize}
\textbf{Overall Score: 3.8} (Quality Level: \textbf{L4})
\end{minipage}%
\begin{minipage}[t]{0.52\columnwidth}
\textbf{Explanation:}\\
The reasoning demonstrates strong organization with a clear three-step approach that systematically addresses the key issues in the problem. The logical flow from addressing the off-target effects to oxidative stress is well-structured, and each step builds upon the previous one. The explanation provides good justification for why other options are inferior, though it could benefit from more explicit self-verification steps.
\end{minipage}

\vspace{0.5em}
\textbf{Strengths:}
\begin{itemize}[leftmargin=*, nosep]
    \item Clear step-by-step organization with numbered steps
    \item Strong justification for eliminating alternative options
    \item Logical progression from immediate concerns to longer-term considerations
    \item Good technical accuracy in explaining why RNAi won't work in E. coli
    \item Effective comparison of the relative merits of each intervention
\end{itemize}

\textbf{Weaknesses:}
\begin{itemize}[leftmargin=*, nosep]
    \item Limited explicit self-verification steps or cross-checking of conclusions
    \item Could provide more quantitative reasoning for why the chosen sequence is optimal
    \item Missing discussion of potential risks or drawbacks of the chosen approach
    \item Limited exploration of how the two steps in option (a) might interact with each other
    \item Could elaborate more on the timeline and implementation details
\end{itemize}

\textbf{Suggestions for Improvement:}
\begin{itemize}[leftmargin=*, nosep]
    \item Add explicit verification steps to confirm the logic at each stage
    \item Include more discussion of potential risks and mitigation strategies
    \item Provide more detailed explanation of why addressing off-target effects should precede oxidative stress treatment
    \item Add specific examples or scenarios to illustrate the expected outcomes
    \item Include a brief discussion of implementation timeline and resource requirements
\end{itemize}
\end{tcolorbox}

This example illustrates how the judge model provides a comprehensive evaluation of reasoning quality across multiple dimensions. The defender's solution received an overall score of 3.8 (L4 quality level), indicating strong reasoning with well-structured organization but with opportunities for improvement in self-verification and detailed implementation considerations.

\end{document}